\documentclass{article}
\usepackage{iclr2026_conference,times}

\usepackage[T1]{fontenc}

\usepackage{amsmath,amsfonts,bm}

\def\eqref#1{equation~\ref{#1}}

\def\1{\bm{1}}

\DeclareMathAlphabet{\mathsfit}{\encodingdefault}{\sfdefault}{m}{sl}
\SetMathAlphabet{\mathsfit}{bold}{\encodingdefault}{\sfdefault}{bx}{n}

\usepackage{hyperref}
\usepackage{url}
\usepackage{natbib}
\usepackage{amsmath, amssymb, amsfonts}
\usepackage{graphicx}
\usepackage{multirow}
\usepackage[normalem]{ulem}
\usepackage[cache=false]{minted}
\usepackage[autolanguage]{numprint}
\usepackage{interval}
\usepackage{diagbox}
\usepackage{tcolorbox} 
\usepackage{makecell}
\usepackage{syntax}
\usepackage{booktabs}   
\usepackage{multirow} 
\usepackage{tikz}
\usepackage{dsfont}
\usepackage{array}

\setminted[json]{
    mathescape,
    linenos,
    numbersep=-4pt,
    gobble=2,
    frame=lines,
    framesep=1mm,
    breaklines,
    fontsize=\footnotesize,
    escapeinside=||
}

\setminted[shell]{
    mathescape,
    linenos,
    numbersep=-4pt,
    gobble=2,
    frame=lines,
    framesep=1mm,
    breaklines,
    fontsize=\footnotesize,
    escapeinside=||
}

\setminted[python]{
    mathescape,
    linenos,
    numbersep=-4pt,
    gobble=2,
    frame=lines,
    framesep=1mm,
    breaklines,
    fontsize=\footnotesize,
    escapeinside=||
}

\setminted[c++]{
    mathescape,
    linenos,
    numbersep=-4pt,
    gobble=2,
    frame=lines,
    framesep=1mm,
    breaklines,
    fontsize=\footnotesize,
    escapeinside=||
}

\newcommand{\toolname}{\textsc{AgentScope}}
\newcommand{\datasetname}{AgentErrata}
\newcommand{\graphname}{Reasoning-Action Graph}
\newcommand{\graphabbr}{ReAG}

\newcommand{\failname}{failure mode}
\newcommand{\failnameup}{Failure Mode}
\newcommand{\failnameupfirst}{Failure mode}

\newcommand{\agentfailxnum}{303}

\title{Diagnosing with Insights: Structured Analysis of Agent Failures via Behavioral Abstractions}

\author{
Jiayi Bi\textsuperscript{1},
Yanjie Gao\textsuperscript{2,$\dagger$},
Yuanmin Xie\textsuperscript{1},
Liqun Li\textsuperscript{3},
Tianyin Xu\textsuperscript{4},
Fan Yang\textsuperscript{2},
Mao Yang\textsuperscript{2}\\[3pt]
\textsuperscript{1}Tsinghua University\\
\textsuperscript{2}Microsoft Research\\
\textsuperscript{3}Microsoft\\
\textsuperscript{4}University of Illinois Urbana-Champaign
}

\iclrfinalcopy 
\begin{document}

\maketitle
\begingroup
\renewcommand{\thefootnote}{}
\footnotetext{This work was done during Jiayi's internship at Microsoft Research and Yuanmin's internship at Microsoft. $^{\dagger}$Corresponding author.}
\addtocounter{footnote}{-1}
\endgroup

\begin{abstract}

With the proliferation of LLM agents, the ability to understand and diagnose failures in agents is essential to achieving superior effectiveness and trustworthiness.
As agent failures often manifest via long and complex trajectories, manually finding the needles in the haystack is untenable. 
However, traditional diagnosis techniques for software bugs can hardly address LLM agent failures, while completely relying on LLMs as the judge yields unreliable diagnosis results. To overcome these challenges, this paper presents \toolname{}, a new neuro-symbolic approach for agent {\failname} diagnosis.
The key principle of \toolname{} is to abstract agent behavior, based on its trajectories, into structured representations.
Furthermore, \toolname{} introduces the concept of neural invariants to specify agent behavior properties.
\toolname{} leverages LLM-guided reasoning atop the structured representation against neural invariants to pinpoint both the failure step and its type in the trajectory.
We show the effectiveness of \toolname{} on publicly available agent failure datasets (Who\&When) and a more comprehensive dataset created by us (\datasetname{}), where
\toolname{} significantly outperforms the current art in fault localization and attribution accuracy.
Our work shows that integrating structured abstractions with LLM-guided reasoning enables effective, reliable, and interpretable diagnosis for agent failures.

\end{abstract}

\section{Introduction}
\label{sec:intro}

Recent advances in Large Language Models (LLMs) have been driving active development of LLM agents that autonomously interact with tools and environments using natural languages.
For example, agents can issue API calls, synthesize code, query databases, and cooperate with other agents to solve complex and multi-step tasks. Modern agent frameworks such as LangChain~\citep{chase2022langchain}, Auto-GPT~\citep{SignificantGravitasAutoGPT}, and OpenManus~\citep{openmanus2025} orchestrate multi-step interactions, allowing agents to operate in open and dynamic environments.  

Despite their exciting capabilities, LLM agents are known to fail in subtle and sophisticated ways~\citep{cemri2025multi,zhang2025agent, 10.5555/3600270.3600957, bryan2025taxonomy, baker2025monitoring, fu2025scaling}. Failures can occur at any step during the agent's {\it reasoning} (e.g., the chain of thoughts) and {\it action execution} (e.g., tool calls), cascade along the agent runtime behavior, and eventually manifest as specific mistake behavior or even disrupt task execution.
Thus, the ability to understand and diagnose failures in agents  is essential to achieving superior effectiveness and trustworthiness of agent systems.
Unfortunately, as agent failures often manifest through prolonged and complex trajectories composed of numerous steps with accumulating context, manually finding needles in the haystack is slow, costly, and obviously untenable.

Worse still, traditional diagnosis techniques for software bugs can hardly address LLM agent failures~\citet{zeller:09,attariyan:10,yuan:10,zhang:19,ren:osdi:23}. 
The fundamental reason is that traditional techniques are confined to symbolic and logical analysis of software codes and program executions.
In contrast, agent failures are often rooted in faulty reasoning errors, invalid contexts, or instruction-unfollowing behaviors across multiple steps, which involve the entanglement of fuzzy neural paradigms and rigid symbolic paradigms.

Recent studies~\citep{zhang2025agent,cemri2025multi,zhu2025raffles,dlshriverintercepts2023} have proposed several neural approaches for agent failure analysis---prompting or fine-tuning LLMs with failure trajectories and asking them to identify root causes.
However, our experiments show that sole LLM-based approaches often produce unreliable and incomplete diagnostic results. 
Even the best-performing model, GPT-5.1~\citep{singh2025openai}, achieves only 18.15\% accuracy on our failure attribution datasets (see Section~\ref{rq:attri}).
The cause of this limitation is that LLMs, even when fine-tuned, struggle to systematically capture multi-step reasoning and action behaviors and maintain consistent causal invariants. Thus, they are prone to confusing correlated symptoms with true causes and sensitive to contexts and instructions.

This paper presents {\toolname}, a novel neuro-symbolic framework for diagnosing agent failures.
The key idea of \toolname{} is to abstract the agents' behavior from their trajectories into a {\it structured} representation, termed the {\graphname} ({\graphabbr}), that encapsulates both the reasoning and action execution steps of the agents, enabling rigorous correctness reasoning using formally defined invariant violation conditions on their behaviors.
Different from traditional program invariants~\citep{hoare:69}, \toolname{} introduces the concept of {\it neural invariants}, which can be specified with neural functions to encode correctness conditions; checking such invariants requires LLM-guided reasoning.
We show that LLM-guided reasoning atop structured behavior abstractions enables \toolname{} to effectively diagnose the causes of the agent failure.
Compared with existing vanilla LLM-as-a-judge approaches,
\toolname{} demonstrates stronger diagnosis ability and more interpretable results.
It can accomplish both {\it failure localization} and {\it failure attribution}: the former pinpoints the root-cause step and the latter predicts the failure category.

We evaluated {\toolname} on a publicly available agent failure dataset named Who\&When~\citep{zhang2025agent} and a new dataset called \datasetname{} created by us through failure-taxonomy-guided fault injection for more comprehensive evaluation.
Results show that {\toolname} achieves significant improvements in accuracy and interpretability over state-of-the-art approaches, with localization accuracies ranging from 25.40\% to 77.78\% on the Who\&When Algorithm-Generated dataset and from 22.41\% to 34.48\% on the Who\&When Hand-Crafted dataset, along with enhanced attribution interpretability. Moreover, {\toolname} also outperforms other methods on our new dataset, \datasetname{}, achieving accuracies ranging from 28.38\% to 54.13\%.

In summary, we make the following contributions:

\begin{itemize}
  \setlength\itemsep{1pt}
  \vspace{-5pt}
  \item {\bf Principle.} We show that LLM-guided reasoning atop structured representations of behavior abstractions
  can significantly improve agent failure diagnosis ability.

  \item {\bf Concept.} We introduce the concept of neural invariants, which use neural functions to formally define and detect properties of agent misbehaviors. 

  \item {\bf Tooling.} We develop {\toolname}, a practical runtime framework and toolchain for agent failure detection and diagnosis that is integrated with modern agent frameworks.

	\item {\bf Dataset.} We create \datasetname{}, a new benchmark dataset for agent failures. Generated through a failure-taxonomy-guided fault injection process, it contains a comprehensive set of failure types covering diverse agent misbehaviors.

  \item {\bf Evaluation.} We present the utility of {\toolname} using systematic evaluations and showcase its improved accuracy and interpretability over the current art.
\end{itemize}

\section{Background}
\label{sec:background}

Agent failures have a variety of causes through incorrect reasoning and action execution steps, in which a single failure step can cascade and lead to observable symptoms.
Table~\ref{tab:error-main} summarizes our taxonomy of agent failure patterns (hereafter referred to as {\failname}s), organized into three dimensions by where the failure manifests in an agent trajectory: \textbf{(i) Reasoning}, \textbf{(ii) Control-flow}, and \textbf{(iii) Action}.
\textbf{Reasoning} refers to failures in internal decision-making and context utilization.
\textbf{Control-flow} captures failures in execution orchestration and control, including step transitions and termination.
\textbf{Action} involves failures in executing decisions through external tools or environments.
Reasoning failures include \textit{Insufficient Context}, \textit{Wrong Context}, \textit{Instruction Unfollowing}, and \textit{Context Miss}.
Control-flow failures capture \textit{Termination Miss}, \textit{Premature Termination}, and \textit{Step Loop}.
Action failures include \textit{Action Mismatch}, \textit{Invocation Issue}, and \textit{Execution Failure}.
This taxonomy provides a comprehensive and interpretable framework for understanding and diagnosing failures of LLM-based agents. The taxonomy is derived based on our empirical analysis of failure trajectories, along with references to existing community categorizations~\citep{cemri2025multi,zhang2025agent,baker2025monitoring}.

\begin{table}[H]
    \vspace{-10pt}
    \centering
    \caption{Taxonomy of agent {\failname}s.}
    \label{tab:error-main}
    \def\arraystretch{1.2}%
    \scalebox{0.735}{
        \begin{tabular}{!{\vrule width 1pt} m{1.8cm} | m{4.2cm} | m{11cm} !{\vrule width 1pt}}
            \hline
            \bf Dimension & \bf Category             & \bf Description                                                                  \\
            \hline
            \hline
            \multirow{6}{*}{\makecell{Reasoning}}
   
                         & Wrong Context (WC)          & Uses context that is irrelevant, invalid, or misleading, resulting in erroneous outcomes. \\
            \cline{2-3}
                         & Instruction Unfollowing (IU) & Deviates from the provided instructions or specifications, producing outputs inconsistent with task requirements.                         \\
            \cline{2-3}
                                  & Insufficient Context (IC)   & Missing essential information in the LLM input causes reasoning and decision errors.  \\
            \cline{2-3}
                         & Context Miss (CM)           & Overlooks relevant information from history, causing incomplete reasoning.         \\
            \hline
            \hline
            \multirow{5}{*}{\makecell{Control-flow}}
                         & Termination Miss (TM)        & Continues executing steps beyond task completion, causing unnecessary or infinite actions.        \\
            \cline{2-3}
                         & Premature Termination (PT)  & Stops execution before the task is fully completed, resulting in incomplete solutions or missing steps.                 \\
            \cline{2-3}
                         & Step Loop (SL)              & Repeats prior steps without justification, leading to redundant or cyclical behavior.                         \\
            \hline
            \hline
            \multirow{4}{*}{\makecell{Action}}
                         & Action Mismatch (AM)        & Selects actions (tools, APIs, etc.) that contradict its reasoning, resulting in misalignment between decision and execution.  \\
            \cline{2-3}
                         & Invocation Issue (II)       & Incorrectly invokes tools or APIs, producing unintended or failed operations.                   \\
            \cline{2-3}
                         & Execution Failure (EF)    & Encounters runtime errors while executing a tool, API call, or other action, preventing successful task completion.                      \\
            \hline
        \end{tabular}
    }
    \vspace{-10pt}
\end{table}


\section{Methodology}
\label{sec:approach}

\subsection{Behavioral Abstraction}

In \toolname{}, an agent system's trajectory is represented as a structured graph, termed {\graphname} ({\graphabbr}). Formally, a {\graphabbr} is a directed acyclic graph:
\[
G = \{V, E\}
\]
where $V$ is the set of vertices representing individual steps, and $E$ is the set of edges encoding control or data dependencies between steps. Each vertex is represented as a quadruple:
\[
v_i = \langle id_i, r_i, c_i, \mathcal{I}_i \rangle
\]
where $id_i$ is the unique step identifier, $r_i$ is the role of the agent performing the step, $c_i$ is the operational content, and $\mathcal{I}_i$ is the \textit{Intermediate Semantic Representation (ISR)} of the step.

To address redundancy and long-context challenges, we construct a quickly analyzable trajectory index and memory, which motivates the design of the ISR.
The ISR for each step can be formally represented as a tuple of three sub-components:
\[
\mathcal{I}_i = (C_i, R_i, S_i)
\]
where $C_i$ denotes the \textit{Intent and Context} component, $R_i$ denotes the \textit{Reasoning and Action} component, and $S_i$ denotes the \textit{Signal and Validation} component. Each sub-component abstracts different semantic aspects of the step and supports downstream modules in interpreting, validating, and reusing information across long agent trajectories. The \textit{Intent and Context} component $C_i$ encodes task goals, instructions, and the purpose of the step. The \textit{Reasoning and Action} component $R_i$ captures the agent's decision-making, reasoning summaries, performed actions, external tool invocations, and outputs. The \textit{Signal and Validation} component $S_i$ encodes internal knowledge usage, confidence, and termination signals. All three ISR components provide structured information to facilitate step identification and characterization.

Edges $e \in E$ represent control or data dependencies between steps, capturing the logical flow of reasoning and execution dynamics. Vertices are constructed through instrumentation of API calls, tool interactions, and system logs, and are further refined with semantic parsing to ensure coherent step boundaries. Even unstructured logs can thus be converted into fully annotated, structured graphs with ISR, enabling efficient reasoning, long-term memory accumulation, and downstream analysis.

\subsection{Neural Invariants}

\toolname{} introduces the concept of {\it neural invariants} to specify the behavioral properties of agent failures, which can be used to verify the correctness of the {\graphabbr} of a given trajectory and to diagnose {\failname}s. Specifically, {\toolname} conducts LLM-guided reasoning to detect violations of neural invariants. When an invariant is violated, {\toolname} identifies the vertex (i.e., the step) in the {\graphabbr} where the failure starts to manifest and classifies the type of the failure. Compared with vanilla LLM-as-a-judge approaches~\citep{zhang2025agent, cemri2025multi}, \toolname{} offers several advantages:
(i) precise localization of the failure step in the {\graphabbr},
(ii) fine-grained classification of the failure type based on invariant categories rather than opaque judgment criteria,
(iii) highly interpretable explanations through invariant violations that explicitly expose the root causes of the failure, and
(iv) more faithful and deterministic results, since the verification relies on predefined invariants rather than stochastic model judgments solely .

To enable effective failure diagnosis, we define a comprehensive set of agent behavior neural invariants based on the taxonomy of agent failures (Table~\ref{tab:error-main}). Each {\failname} is formalized as an invariant violation; Appendix~\ref{app:invar} lists all invariant violations used in \toolname{}. The framework can be easily expanded to new {\failname} taxonomies by adding corresponding invariant violations.
Unlike traditional program invariants~\citep{hoare:69}, which rely solely on symbolic conditions, \toolname{} supports semantic conditions via neural functions. Each neural function is implemented as a call to a general-purpose LLM (e.g., GPT-4o) with structured, task-specific prompts that incorporate \graphabbr{} information, enabling reasoning over richer context beyond raw step content. These functions are modular and can be customized, such as by using fine-tuned models.
We illustrate an invariant violation with the following example:

\begin{quote}\footnotesize
	\textbf{Action Mismatch.}
	This failure occurs when an assistant's action contradicts or fails to align with the intent of the preceding reasoning step.
	We use $iv_f$ to denote the invariant violation associated with failure mode $f$; $iv_f=\mathrm{true}$ indicates that the corresponding neural invariant is violated.
	Let $\mathbf{N}_{action} \subseteq \mathbf{N}$ denote action nodes (tool calls or outputs), and $\mathbf{N}_{reason} \subseteq \mathbf{N}$ denote reasoning nodes.
	For an action node $n_t$ and its preceding reasoning node $n_{t-1}$, define
	$\texttt{aligned}(purpose_{t-1}, action_t, tool\_name_t, tool\_args_t)$
	to check whether the action type matches the intent, the tool choice is appropriate, and the action advances the task.
	The invariant violation is expressed as:
\begin{multline*} 
	iv_{mismatch} ::= \exists~n_t \in \mathbf{N}_{action},~n_{t-1} \in \mathbf{N}_{reason} : \\
	\text{\(\neg\, \texttt{aligned}(purpose_{t-1},\; action_t,\; tool\_name_t,\; tool\_args_t)\)} \\
\end{multline*}
	
	A violation indicates misalignment between the action and its stated purpose (e.g., tool misuse, missing or incorrect actions, or hallucinated progress).
	When implemented with an LLM judge, $\texttt{aligned()}$ is treated as a binary classifier.
\end{quote}

\begin{figure}[t]
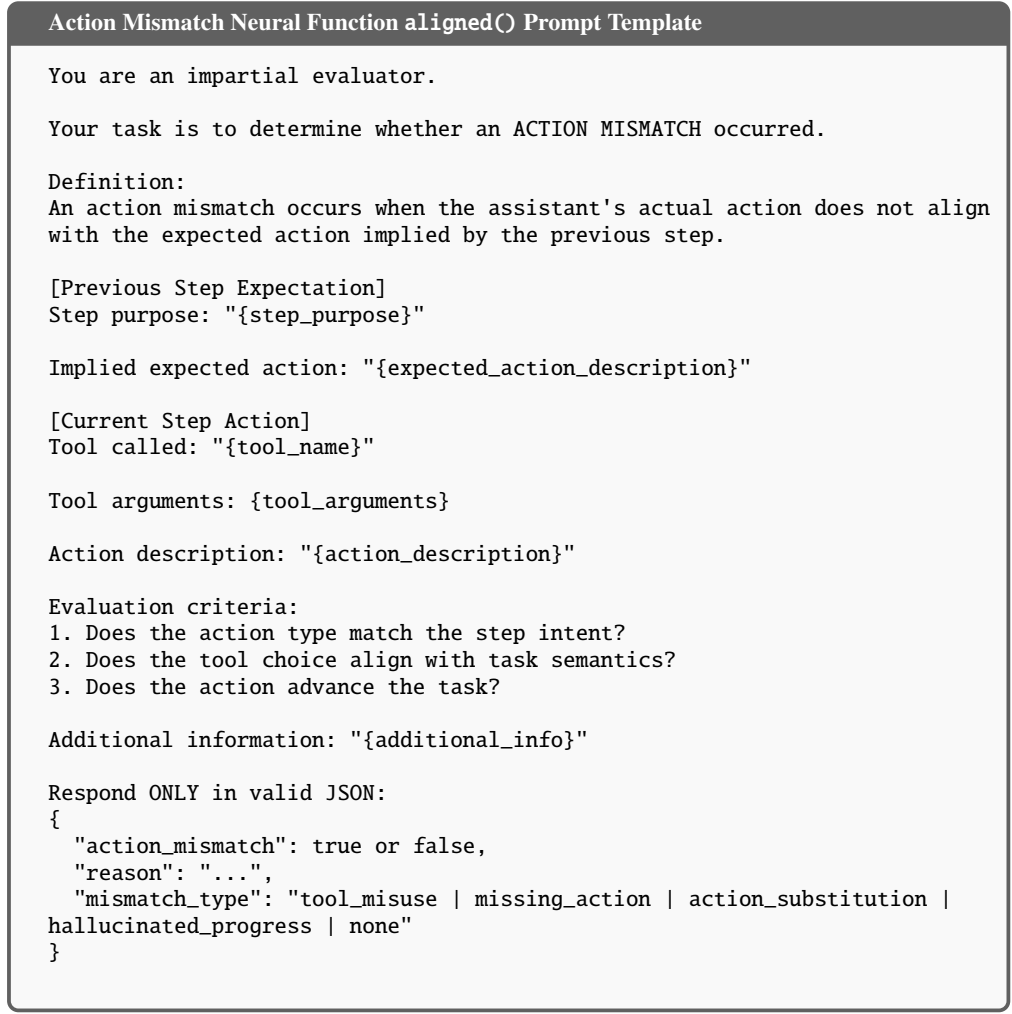

	\centering
	\begin{minipage}{0.95\linewidth}
		\begin{tcolorbox}[
				colback=gray!5!white,
				colframe=gray!75!black,
				title=Action Mismatch Neural Function \texttt{aligned()} Prompt Template,
				fonttitle=\small\bfseries,
				fontupper=\footnotesize
			]
			\begin{verbatim}
You are an impartial evaluator.

Your task is to determine whether an ACTION MISMATCH occurred.

Definition:
An action mismatch occurs when the assistant's actual action does not align
with the expected action implied by the previous step.

[Previous Step Expectation]
Step purpose: "{step_purpose}"

Implied expected action: "{expected_action_description}"

[Current Step Action]
Tool called: "{tool_name}"

Tool arguments: {tool_arguments}

Action description: "{action_description}"

Evaluation criteria:
1. Does the action type match the step intent?
2. Does the tool choice align with task semantics?
3. Does the action advance the task?

Additional information: "{additional_info}"

Respond ONLY in valid JSON:
{
  "action_mismatch": true or false,
  "reason": "...",
  "mismatch_type": "tool_misuse | missing_action | action_substitution | 
hallucinated_progress | none"
}
			\end{verbatim}
		\end{tcolorbox}
	\end{minipage}
\caption{Prompt template for the action mismatch neural function \texttt{aligned()}. Placeholders (e.g., \texttt{{step\_purpose}}, \texttt{{tool\_name}}) are instantiated from the agent trajectory.}
	\label{fig:action-mismatch-prompt}
\end{figure}

Figure~\ref{fig:action-mismatch-prompt} shows the prompt used by the neural function \texttt{aligned()} in this invariant violation.
Based on the {\failname} taxonomy and corresponding invariant violation, we synthesize diverse test cases across domains to tune and validate the neural functions. We further perform manual verification to ensure their effectiveness.

\subsection{Implementation}
\label{sec:impl}

\begin{figure*}
	\centering
	\includegraphics[width=\textwidth]{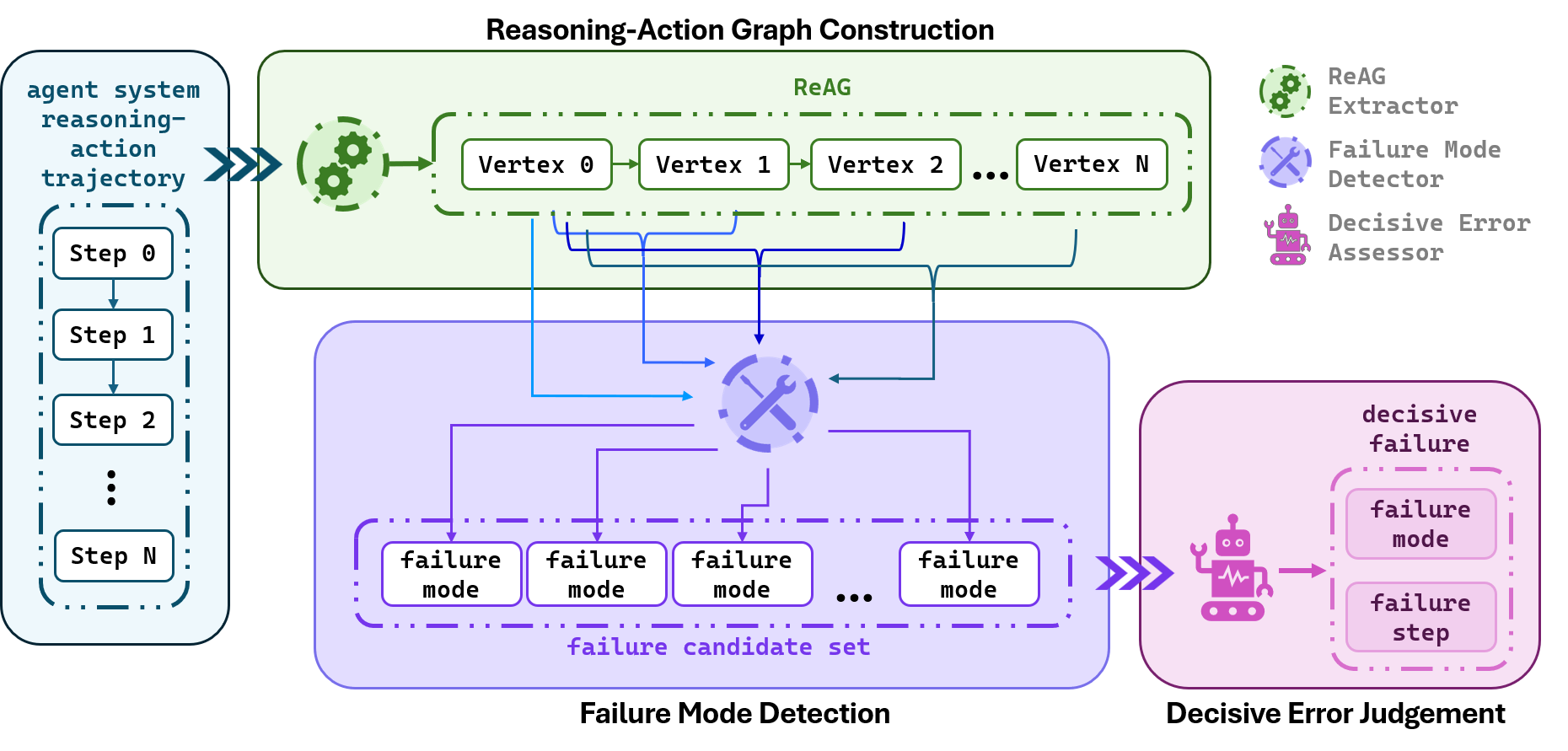}
	\caption{The workflow of {\toolname}.}
	\label{fig:overview}
\end{figure*}

{\toolname} is implemented as a modular framework for online and offline diagnosis of single-agent and multi-agent trajectories. Diagnosing raw agent traces in one shot is inherently unreliable due to the limitations of large language models (LLMs): they struggle with long contexts, are sensitive to token positions, and can lose or misinterpret implicit dependencies in sequential data. To address these challenges, {\toolname} is deliberately structured into three stages, as illustrated in Figure~\ref{fig:overview}. Each stage provides a controlled representation or analysis that reduces the cognitive burden on the model while improving reliability, interpretability, and extensibility. The three stages are as follows:

\textbf{{\graphabbr} construction} converts runtime traces into a unified, annotated representation. Instrumentation captures each step, while LLMs are used selectively to parse reasoning or implicit actions. Each step is enriched with structured semantic fields---such as role, instructions, actions, missing-information signals, and step purpose---to create a stable, aligned representation. This is essential because raw traces alone are brittle: LLMs can misalign steps, forget early context, or misinterpret implicit reasoning. {\graphabbr} ensures subsequent analyses operate on explicitly aligned, semantically meaningful fields rather than raw text.

\textbf{{\failnameupfirst} detection} applies a set of neural invariant checks over the annotated graph. Each check identifies the relevant trajectory steps and context and evaluates them using a task-specific judge respectively, since different failure modes may require evidence from different context ranges and trajectory locations. Candidate failures are then retained with supporting evidence rather than immediately reduced to a single diagnosis, allowing multiple hypotheses to be maintained for broader coverage and auditability while avoiding premature selection.

\textbf{Decisive error judgment} selects a single root-cause failure from the candidate set using the full trajectory and optional task context. The decisive error is defined not as the first anomaly, but as the failure whose downstream impact most strongly explains the degraded outcome. This stage ensures that correlated or cascading violations are correctly attributed, producing a representative verdict that remains traceable to the evidence surfaced in prior stages. The strategy could be extended to identify the first failure step or the point beyond which recovery is no longer possible.

Overall, this staged design balances model-driven reasoning with structured analysis. By systematically structuring trajectories, isolating local failures, and performing controlled aggregation, the system mitigates inherent LLM limitations---context fragmentation, positional sensitivity, and retrieval/reasoning errors---while providing a faithful, inspectable, and extensible framework for trajectory diagnosis.

\section{Experiments}
\label{sec:meth}

We evaluated \toolname{} using both proprietary and open-source models, including GPT-4o (version: 2024-11-20)~\citep{hurst2024gpt}, GPT-5.1 (version: 2025-11-13)~\citep{liu2024deepseek}, and DeepSeek-V3.2~\citep{yang2025qwen3}. All models were deployed via Azure AI Foundry~\citep{azureaifoundry}. We used a consistent hyperparameter configuration across all models, with $temperature = 0.01$ and $max\_tokens = 4096$, while keeping all other hyperparameters at their default values.

\subsection{Benchmarks}
We used two benchmark datasets for evaluation.

{\bf Who\&When.} The Who\&When dataset~\citep{zhang2025agent} targets failure localization.
It evaluates agents on the understanding towards temporal and causal relations in narrative contexts.
It contains 184 human-annotated trajectories, each involving identifying who did what and when, emphasizing temporal alignment and attribution across multiple events.
The Who\&When dataset includes two sub-datasets that are generated from algorithm generated and hand-crafted agent systems respectively.
We denote them as ``Algo-Generated'' and ``Hand-Crafted'' in \S\ref{sec:results}
and evaluate the two sub-datasets separately. The former contains 126 traces and the latter contains 58 traces.

{\bf \datasetname{}.}
The Who\&When dataset annotates failure steps but lacks explicit failure type labels, limiting its usefulness for evaluating failure attribution. It typically marks the first step where a potential mistake arises~\cite{zhang2025agent}. However, as models become more fault-tolerant, early failures may be corrected by powerful agents' self-repair and therefore may not lead to final task failure (see case study in Appendix~\ref{whonwhen}). This can overemphasize minor mistakes while overlooking critical failures, such as repeated instructions or missing information. To address this, we construct \datasetname{}, which labels both failure steps and types and ensures that injected errors result in task failure. By using failure-taxonomy-guided fault injection, \datasetname{} captures diverse failure patterns, enabling rigorous evaluation of debugging tools and highlighting weaknesses in failure detection.
We select trajectories from agent frameworks and benchmark datasets represented in the current release of \datasetname{}. 
For agent frameworks, we include OpenManus~\citep{openmanus2025}, OWL~\citep{hu2025owl}, and Mini SWE Agent~\citep{yang2024sweagent}, which span complementary paradigms of LLM-based agents, ranging from tool augmented general purpose workflows, to reasoning centric autonomous agents, and domain-specialized software engineering agents.
For benchmark datasets, we draw trajectories from BrowseComp~\citep{wei2025browsecompsimplechallengingbenchmark}, SWE-Bench Lite~\citep{jimenez2024swebenchlanguagemodelsresolve}, and WebArena~\citep{zhou2024webarenarealisticwebenvironment}, which span information-seeking web browsing, real-world software engineering problem solving, and interactive web environment tasks.
To create \datasetname{}, we first configure the selected agents to generate execution trajectories and then filter out only those that successfully complete the tasks, thereby minimizing the impact of unlabeled and irrelevant erroneous trajectories in the dataset.
We then inject failures into the original agent executions that trigger issues such as instruction unfollowing, step loops, and premature termination. These failures not only cause individual step errors but also prevent the overall task from being completed successfully, which we refer to as decisive errors. For failures that can be injected at individual steps, we employ instrumented replay to inject them at designated steps. For cases where failures involve cross-step semantic and textual dependencies, we further apply post-processing to the trajectories to identify and align such inconsistencies.
Injections are applied across all categories in the current taxonomy (see Table~\ref{tab:error-main}) with a rough uniform distribution. To maintain contextual coherence, we additionally leverage an LLM to assess whether a given trajectory is suitable for specific categories of \failname{} before the failure injection. 
To further ensure the quality of the dataset, all injected trajectories are manually verified, and any invalid or unreasonable trajectories are discarded.
{\datasetname} currently contains {\agentfailxnum} high-quality trajectories, providing a comprehensive dataset for evaluating diagnosability of agent failures. We show the detailed distribution of {\datasetname} in Appendix~\ref{app:datadis}.

We did not use the MAST dataset~\citep{cemri2025multi}, as it only provides annotations of overall failure types but lacks specific failure steps and underlying reasons. Consequently, it cannot be directly employed to reliably analyze the root causes of misattributions or to automatically evaluate the limitations of the benchmarked tools.

\subsection{Evaluation Metrics}

We use the two metrics from~\citet{zhang2025agent} and introduce an additional metric to measure the effectiveness of diagnosis:

\begin{quote}\small

    {\bf Step-level accuracy (SLA)} quantifies the percentage of correctly identified root-cause failure steps. Formally, let \( N \) be the total number of samples, \( S_i \) be the true root-cause failure step for sample \( i \) and \( \hat{S}_i \) the predicted step, then
    $\text{SLA} = \frac{1}{N} \sum_{i=1}^N \mathds{1}\left(\hat{S}_i = S_i \right)$, where \(\mathds{1}(\cdot)\) is the indicator function.

    {\bf Step-level accuracy with tolerance (SLAT)} considers a prediction correct if the predicted step falls within a tolerance range \(\delta\) of the actual root-cause step: $\text{SLAT} = \frac{1}{N} \sum_{i=1}^N \mathds{1}\left(|\hat{S}_i - S_i| \leq \delta \right)$.
    In other words, SLAT tolerates slight inaccuracy.

    {\bf Classification accuracy (CA)} stands for the percentage of correctly classified \failname{}s. Let $M_i$ be the ground truth \failname{} for the $i^{th} $ trace and $\hat{\mathrm{M}}_i = \{\hat{M}_i^{(l)}\}_{l = 1}^{L}$ be the predicted \failname{}s, where $L$ is the total number of predicted \failname{}s, then CA is defined as $\text{CA} = \frac{1}{N}\sum_{i=1}^N \mathds{1} \left(\exists\hat{M}_i^{(l \in [1, L], l \in \mathds{Z})} \in \hat{\mathrm{M}}_i, \hat{M}_i^{(l)} = M_i\right)$.

\end{quote}

The Agent-level accuracy (ALA) from ~\citet{zhang2025agent} is not adopted because step-level metrics already reflect the responsible agent, and knowing the agent role alone does not indicate the exact failure step or cause, while agent roles are often skewed---guessing frequent agents can yield high ALA without reflecting true diagnosis ability. Thus, ALA provides little additional insight.

\section{Results}
\label{sec:results}

\subsection{How effective is {\toolname} in failure localization?}

To evaluate the effectiveness of {\toolname} in localizing the root-cause failure step, we compare it against two representative baseline approaches:

\textbf{\textsc{All-at-once}~\citep{zhang2025agent}}: This approach considers the agent’s entire trajectory as a single unit and uses an LLM to directly identify failures in a single pass. While efficient, it performs holistic reasoning over the full trajectory without intermediate structured decomposition, explicit step alignment, or fine-grained attribution of responsibility across steps. As a result, failure reasoning may suffer from global entanglement of evidence, making it difficult to disentangle which specific step is causally responsible when multiple interacting errors exist.

\textbf{\textsc{Step-by-step}~\citep{zhang2025agent}}: This approach analyzes the trajectory in a sequential manner, where each step is evaluated together with all its preceding steps as context (i.e., prefix-conditioned reasoning). At each position, the model determines whether the current step constitutes an anomaly given the accumulated history. While this introduces local historical context, it still lacks explicit structured representation of the trajectory and does not perform graph-based aggregation or invariant-guided filtering. Consequently, it cannot maintain globally consistent dependencies across the full trajectory, and failure signals are derived from prefix-local judgments rather than global causal attribution over the entire process.

Table~\ref{tab:fault-localization-accuracy4-combined} presents the failure localization accuracy of {\toolname} compared with the baseline methods across three datasets under different tolerance levels.
On the \datasetname{} dataset, {\toolname} achieves dominant performance regardless of whether the ground-truth solution is provided. With GPT-4o (w/o Solution), {\toolname} attains 30.03\% at T$\pm$0, far exceeding W\&W \textsc{Step-by-step} (8.91\%) and \textsc{All-at-once} (2.09\%), and the margin further increases at T$\pm$3, where {\toolname} reaches 49.83\% versus 25.74\% and 23.69\% for the baselines.
With GPT-5.1, {\toolname} still maintains strong performance at 31.35\% (T$\pm$0, w/o Solution), compared to \textsc{Step-by-step} at 1.32\% and \textsc{All-at-once} at 3.30\%. At T$\pm$3, it reaches 54.13\%, versus 13.20\% and 18.81\%, respectively. 
Interestingly, GPT-5.1 does not consistently improve failure-localization accuracy across methods. This suggests that general-purpose model capability alone does not determine diagnostic accuracy. At the same time, {\toolname}'s advantage on \datasetname{} persists across both backbone models, consistent with the benchmark's focus on concrete failures that decisively affect task completion. Further results and analysis are provided in Section~\ref{subsec:rq3}.

\begin{table*}[t]
	\centering
	\caption{SLAT (\%) of the baseline methods and \toolname{} across backbone models and benchmarks; higher values are better.}
	\label{tab:fault-localization-accuracy4-combined}
	\vspace{5pt}
	\resizebox{\textwidth}{!}{
		\begin{tabular}{clccccccccc}
			\toprule
			\multirow{2}{*}{\textbf{Backbone}} & \multirow{2}{*}{\textbf{Method}}
			& \multicolumn{3}{c}{\textbf{\datasetname}} & \multicolumn{3}{c}{\textbf{W\&W Algo-Generated}} & \multicolumn{3}{c}{\textbf{W\&W Hand-Crafted}} \\
			\cmidrule(lr){3-5}\cmidrule(lr){6-8}\cmidrule(lr){9-11}
			& & T$\pm$0 & T$\pm$1 & T$\pm$3 & T$\pm$0 & T$\pm$1 & T$\pm$3 & T$\pm$0 & T$\pm$1 & T$\pm$3 \\
			\midrule
			\multicolumn{11}{l}{\textit{Panel A: w/ Solution}} \\
			\midrule
			\multirow{3}{*}{GPT-4o}
			& W\&W All-at-once & 2.14 & 6.05 & 21.00 & 14.29 & 43.65 & 69.05 & 5.26 & 8.77 & 29.82 \\
			& W\&W Step-by-step & 10.56 & 14.52 & 25.74 & 23.02 & \textbf{53.97} & 73.81 & 15.52 & 15.52 & 20.69 \\
			& \textbf{\toolname} & \textbf{28.38} & \textbf{34.98} & \textbf{48.51} & \textbf{31.75} & 52.38 & \textbf{77.78} & \textbf{25.86} & \textbf{29.31} & \textbf{34.48} \\
			\cmidrule(lr){1-11}
			\multirow{3}{*}{GPT-5.1}
			& W\&W All-at-once & 2.64 & 5.61 & 18.81 & 23.81 & 48.41 & \textbf{79.37} & 3.45 & 13.79 & 25.86 \\
			& W\&W Step-by-step & 1.65 & 4.62 & 12.21 & \textbf{26.19} & \textbf{50.79} & 69.84 & 20.69 & \textbf{31.03} & \textbf{41.38} \\
			& \textbf{\toolname} & \textbf{29.70} & \textbf{34.65} & \textbf{50.50} & 25.40 & 44.44 & 64.29 & \textbf{22.41} & 24.14 & 29.31 \\
			\midrule
			\multicolumn{11}{l}{\textit{Panel B: w/o Solution}} \\
			\midrule
			\multirow{3}{*}{GPT-4o}
			& W\&W All-at-once & 2.09 & 4.18 & 23.69 & 16.67 & 38.89 & 69.84 & 5.17 & 10.34 & \textbf{36.20} \\
			& W\&W Step-by-step & 8.91 & 13.20 & 25.74 & 15.08 & 42.06 & 61.11 & 17.24 & 17.24 & 24.14 \\
			& \textbf{\toolname} & \textbf{30.03} & \textbf{35.64} & \textbf{49.83} & \textbf{28.57} & \textbf{55.56} & \textbf{77.78} & \textbf{22.41} & \textbf{24.14} & 32.76 \\
			\cmidrule(lr){1-11}
			\multirow{3}{*}{GPT-5.1}
			& W\&W All-at-once & 3.30 & 4.62 & 18.81 & 21.43 & 46.83 & \textbf{78.57} & 3.45 & 12.07 & 29.31 \\
			& W\&W Step-by-step & 1.32 & 4.29 & 13.20 & \textbf{25.40} & \textbf{48.41} & 69.05 & \textbf{24.14} & \textbf{32.76} & \textbf{44.83} \\
			& \textbf{\toolname} & \textbf{31.35} & \textbf{36.63} & \textbf{54.13} & \textbf{25.40} & 45.24 & 65.87 & 22.41 & 24.14 & 29.31 \\
			\bottomrule
		\end{tabular}
	}
	\vspace{2pt}

	\footnotesize\centering
	(``W\&W'' denotes ``Who\&When''; ``w/ Solution'' indicates that the ground-truth solution to the original task was provided during analysis, whereas ``w/o Solution'' indicates that it was not provided~\citep{zhang2025agent}.)
\end{table*}

On the Who\&When datasets, {\toolname} with GPT-4o generally outperforms both baselines. For instance, on Who\&When Hand-Crafted (w/ Solution), {\toolname} achieves 25.86\% at T$\pm$0 and 34.48\% at T$\pm$3, substantially surpassing both W\&W \textsc{Step-by-step} (15.52\%, 20.69\%) and W\&W \textsc{All-at-once} (5.26\%, 29.82\%). When the ground-truth solution is not provided, {\toolname} remains robust across both \datasetname{} and Who\&When Hand-Crafted, indicating that the structured graph representation and invariant-guided candidate preservation improve diagnostic stability under missing-context conditions.
However, with GPT-5.1 on Who\&When datasets, {\toolname}'s exact accuracy remains close to the strongest baseline. For example, Who\&When \textsc{Step-by-step} and {\toolname} both reach 25.40\% on Who\&When Algo-Generated (T$\pm$0, w/o Solution). This difference is therefore related to the benchmark's annotation policy rather than a uniform reduction in diagnostic effectiveness under GPT-5.1.
Further inspection of the dataset reveals that some trajectories contain multiple plausible failure points. A cascading failure may include its onset, an observable manifestation, and the point at which it determines the final outcome. Who\&When generally annotates the first mistake, whereas the decisive error judgment stage of {\toolname} selects the candidate that most strongly explains the degraded outcome. Many reviewed GPT-5.1 predictions identify a later verification, answer-submission, or termination step on the same failure-propagation chain as the annotated onset. GPT-4o more often agrees with the earlier reference point in these comparisons, making the distinction more visible under GPT-5.1. A more detailed explanation of the case study is provided in Appendix~\ref{whonwhen}.

Overall, these results suggest that failure localization is not merely a step-wise anomaly detection problem, but a structured multi-stage inference task. {\toolname} decomposes this task into (i) graph-based trajectory construction, (ii) invariant-guided failure-mode detection, and (iii) decisive root-cause selection. First, this design transforms open-ended diagnosis over a raw trajectory into failure-mode-specific checks over relevant steps and context. Compared with \textsc{All-at-once} and \textsc{Step-by-step} reasoning, it reduces the entanglement of global evidence and moves beyond prefix-local analysis. Each detected failure is also linked to localized evidence. The substantial gains on \datasetname{} across both backbone models are consistent with these benefits.
Second, {\toolname} retains multiple detected failure modes before final selection, thereby avoiding premature commitment and separating failure detection from final attribution. The final stage could thus be extended to support different user goals, such as identifying the failure onset, a downstream manifestation, the point beyond which recovery is no longer possible, the final outcome-commitment step, or multiple steps along the failure-propagation chain. Our experiments use the current selection strategy, which prioritizes the failure with the greatest impact on the final outcome. Alternative selection strategies remain future work.

\subsection{How effective is {\toolname} in failure attribution?} \label{rq:attri}

We further evaluate {\toolname}’s ability to classify \failname{}s beyond localizing them, a process referred to as {\it attribution}.
Failure modes in LLM agents can manifest in diverse forms (Table~\ref{tab:error-main}). Existing approaches~\citep{cemri2025multi,zhang2025agent} cannot simultaneously identify the failure step and classify it.
We categorize failures into representative classes and assess {\toolname}’s ability to attribute them within these categories using the {\datasetname} dataset.

Table~\ref{tab:invariantagentfailx-combined} shows that {\toolname} substantially improves failure-attribution accuracy on {\datasetname} compared to the vanilla LLM-as-Judge baseline, which is adapted from the \textsc{All-at-once} approach~\citep{zhang2025agent} and incorporates {\failname} taxonomy information in the prompt.

\vspace{-6pt}
\begin{table}[H]
	\centering
	\caption{Failure attribution on the {\datasetname} dataset (\%) across GPT-4o and GPT-5.1.}
	\label{tab:invariantagentfailx-combined}
	\scalebox{0.85}{
		\begin{tabular}{lcccc}
			\toprule
			\textbf{Model} & \textbf{Method}                  & \textbf{SLA} & \textbf{CA} \\
			\midrule
			GPT-4o         & LLM-as-Judge (w/o Solution)       & 2.80  & 20.63 \\
			& LLM-as-Judge (w/ Solution)        & 2.14  & 19.22 \\
			& \textbf{\toolname} (w/o Solution) & 30.03 & 43.56 \\
			& \textbf{\toolname} (w/ Solution)  & 28.38 & 41.58 \\
			\midrule
			GPT-5.1       & LLM-as-Judge (w/o Solution)       & 3.63  & 18.15 \\
			& LLM-as-Judge (w/ Solution)        & 1.98  & 19.14 \\
			& \textbf{\toolname} (w/o Solution) & 31.35 & 45.87 \\
			& \textbf{\toolname} (w/ Solution)  & 29.70 & 44.88 \\
			\bottomrule
		\end{tabular}
	}
	\\[2pt]
	\begin{minipage}{0.85\linewidth}
		\raggedright\footnotesize
		(``w/ Solution'' indicates that the ground truth solution to the original problem that the agents were solving was given in the context during analysis, whereas ``w/o Solution'' denotes no ground truth solution was given~\citep{zhang2025agent}.)
	\end{minipage}
\end{table}
\vspace{-6pt}

Beyond raw performance gains, these results suggest that failure attribution is fundamentally a \emph{structural reasoning problem} rather than a knowledge or labeling problem. In particular, the large improvement in Step-Level Accuracy (SLA), where {\toolname} achieves 28.38--31.35\% compared to only 1.98--3.63\% for the baseline (a 8--16$\times$ gain), indicates that monolithic LLM-as-Judge methods suffer from severe step misalignment in long-horizon trajectories. This is largely due to positional sensitivity and context compression effects, which make it difficult to consistently identify the true temporal locus of failure when reasoning is performed in a single pass over raw traces.

In contrast, {\toolname} explicitly addresses this limitation through its staged design. The {\graphabbr} construction stage reduces representational ambiguity by transforming raw trajectories into semantically aligned, step-wise structured graphs, mitigating positional bias and enabling consistent cross-step reference. On top of this, the {\failname} detection stage performs localized invariant checks over the structured representation, producing multiple candidate failure hypotheses instead of collapsing them into a single unstable judgment. This decomposition significantly improves recall in complex trajectories where failures may be latent, distributed, or causally entangled.

The improvement in Classification Accuracy (CA), which more than doubles from 18.15--20.63\% to 41.58--45.87\%, further indicates that accurate attribution requires not only correct localization but also reliable disambiguation among competing failure modes. Here, the decisive error judgment stage plays a critical role by aggregating global trajectory evidence to resolve competing hypotheses, distinguishing root causes from downstream symptoms, and preventing over-attribution to superficial anomalies.
Notably, performance remains stable regardless of whether ground-truth solutions are provided, suggesting that the gains are not driven by task leakage or solution conditioning, but by structural decomposition of the attribution process itself.

\subsection{How does {\toolname} perform with different base models?}
\label{subsec:rq3}

This research question investigates the influence of building {\toolname} on different base LLMs. Since LLM-based agents can vary significantly in model type and size, it is important to evaluate whether {\toolname} maintains high failure localization accuracy and attribution performance across these variations.
We conduct experiments using representative LLMs, including GPT-4o, GPT-5.1, and DeepSeek-V3.2. For each base model, we measure {\toolname}'s SLA and CA in detecting {\failname}, and compare the results to baseline methods. Evaluations are performed on both the Who\&When and \datasetname{} benchmarks.

\begin{table}[H]
    \centering
    \caption{{\failnameupfirst} attribution across different models on Who\&When and {\datasetname}. (\%) }

    \label{tab:rq32robustness}
    \scalebox{0.73}{
        \begin{tabular}{llcccccccccccc}
            \toprule
            \multirow{2}{*}{\textbf{Model}} & \multirow{2}{*}{\textbf{Method}} & \multicolumn{2}{c}{\textbf{W\&W A-G}} & \multicolumn{2}{c}{\textbf{W\&W H-C}} & \multicolumn{4}{c}{\textbf{\datasetname}} \\
            \cmidrule(lr){3-4} \cmidrule(lr){5-6}\cmidrule(lr){7-10} & & \textbf{SLA w/o} & \textbf{SLA w/} & \textbf{SLA w/o} & \textbf{SLA w/} & \textbf{SLA w/o} & \textbf{CA w/o} & \textbf{SLA w/} & \textbf{CA w/} \\
            \midrule
            GPT-4o & W\&W All-at-once & \underline{16.67} & 14.29 & 5.17 & 5.26 & 2.09 & -- & 2.14 & -- \\
             & W\&W Step-by-step & 15.08 & \underline{23.02} & \underline{17.24} & \underline{15.52} & \underline{8.91} & -- & \underline{10.56} & -- \\
             & \textbf{{\toolname }} & \textbf{28.57} & \textbf{31.75} & \textbf{22.41} & \textbf{25.86} & \textbf{30.03} & \textbf{43.56} & \textbf{28.38} & \textbf{41.58} \\

            \midrule
            GPT-5.1 & W\&W All-at-once & 21.43 & 23.81 & 3.45 & 3.45 & \underline{3.30} & -- & \underline{2.64} & -- \\
            & W\&W Step-by-step & \textbf{25.40} & \textbf{26.19} & \textbf{24.14} & \underline{20.69} & 1.32 & -- & 1.65 & -- \\
             & \textbf{{\toolname}} & \textbf{25.40} & \underline{25.40} & \underline{22.41} & \textbf{22.41} & \textbf{31.35} & \textbf{45.87} & \textbf{29.70} & \textbf{44.88} \\

            \midrule
            DeepSeek-V3.2 & W\&W All-at-once & 26.98 & 25.40 & 3.57 & 3.51 & 0.66 & -- & 0.99 & -- \\
             & W\&W Step-by-step & \underline{34.13} & \underline{36.51} & \underline{15.51} & \underline{17.24} & \underline{17.82} & -- & \underline{17.82} & -- \\
             & \textbf{{\toolname}} & \textbf{36.51} & \textbf{38.10} & \textbf{25.86} & \textbf{24.14} & \textbf{34.98} & \textbf{45.21} & \textbf{33.33} & \textbf{44.22} \\
            \bottomrule
        \end{tabular}
    }
    \\[2pt]
    \begin{minipage}{\linewidth}
        \raggedright\footnotesize
        (``-'' indicates that the current tool does not support this detection feature; ``W\&W'' stands for ``Who\&When''; ``A-G'' stands for ``Algo-Generated''; ``H-C'' stands for ``Hand-Crafted''; ``w/ '' indicates that the ground truth solution to the original problem that the agents were solving was given in the context during analysis, whereas ``w/o'' denotes no ground truth solution was given~\citep{zhang2025agent}.)
    \end{minipage}

\end{table}

As shown in Table~\ref{tab:rq32robustness}, {\toolname} demonstrates consistently strong performance across all evaluated base models, while maintaining notably low variance compared to baseline methods.

On \datasetname{}, {\toolname} achieves SLA scores of 30.03\% (GPT-4o), 31.35\% (GPT-5.1), and 34.98\% (DeepSeek-V3.2) under the w/o setting, resulting in a range of 4.95 percentage points. This relatively stable performance suggests that {\toolname} effectively decouples failure localization from base model idiosyncrasies. In contrast, baseline methods exhibit strong model sensitivity: \textsc{Step-by-step} drops from 8.91\% (GPT-4o) to 1.32\% (GPT-5.1), indicating that sequential reasoning-based diagnosis is vulnerable to shifts in instruction-following behavior and internal reasoning styles across models. \textsc{All-at-once} remains consistently weak across all models, further suggesting that compressing long-horizon trajectories into a single inference step significantly limits robustness.

A key observation is that {\toolname} not only improves absolute performance but also enhances cross-model invariance. Compared to the strongest baseline per model, {\toolname} yields substantial gains in SLA on \datasetname{}, demonstrating consistent improvements regardless of backbone architecture. This indicates that the proposed structured pipeline is largely independent of model-specific reasoning biases, instead relying on explicit trajectory analysis and invariant-based checking.
A more critical insight lies in the relationship between SLA and CA. While SLA measures whether a failure is correctly localized, CA evaluates whether the root cause is correctly attributed. {\toolname} exhibits a relatively stable and consistently high SLA--–CA alignment across all models (e.g., 30.03\% SLA vs. 43.56\% CA on GPT-4o), suggesting that once a failure is detected, its causal attribution can also be reliably recovered. This implies that the intermediate representations (ReAG construction and ISR-style annotations) preserve causal signals throughout the pipeline, reducing information loss during aggregation and preventing early-stage mislocalizations from propagating into final judgments.

On the Who\&When benchmark, performance exhibits a clearer structure-dependent separation. On the Hand-Crafted (H-C) subset, {\toolname} achieves 22.41\%--25.86\% SLA (w/o), substantially outperforming \textsc{All-at-once} (3.45\%--5.17\%) and remaining competitive with \textsc{Step-by-step} (15.51\%--24.14\%). This suggests that trajectories in the Hand-Crafted subset exhibit deeper compositional dependencies and longer-range implicit interactions, with errors arising across multiple coupled steps rather than at isolated points.
In such settings, naive aggregation or purely sequential reasoning fails to capture cross-step dependencies, while structured decomposition remains effective.

In contrast, the Algo-Generated (A-G) subset yields higher scores and smaller performance differences across methods.
Trajectories in the Algo-Generated subset are generally shorter, which may make relevant failure evidence easier to isolate and partly explain the higher scores across all methods. Nevertheless, \toolname{} performs well on both the Algo-Generated and Hand-Crafted subsets.
An additional finding is that failure diagnosis is not monotonically correlated with model capability. Although GPT-5.1 is generally stronger in generation tasks, it does not consistently outperform GPT-4o or DeepSeek-V3.2 in diagnostic accuracy under baseline methods. 
This suggests that general-purpose model capability alone does not determine diagnostic accuracy; how failure evidence is interpreted and prioritized also matters.

Overall, these results indicate that \toolname{} achieves higher diagnostic accuracy than the evaluated baselines and maintains strong performance across different LLM backbones. By combining ReAG-based reconstruction with ISR-guided invariant checking, \toolname{} organizes trajectory evidence into structured, localized failure modes before final attribution. This design makes the diagnostic process more transparent and auditable across heterogeneous LLM backbones.

\subsection{What is the runtime overhead of {\toolname}?}

\begin{figure}[H]
	\centering
	\begin{minipage}[b]{0.48\textwidth}
		\centering
		\includegraphics[height=4cm]{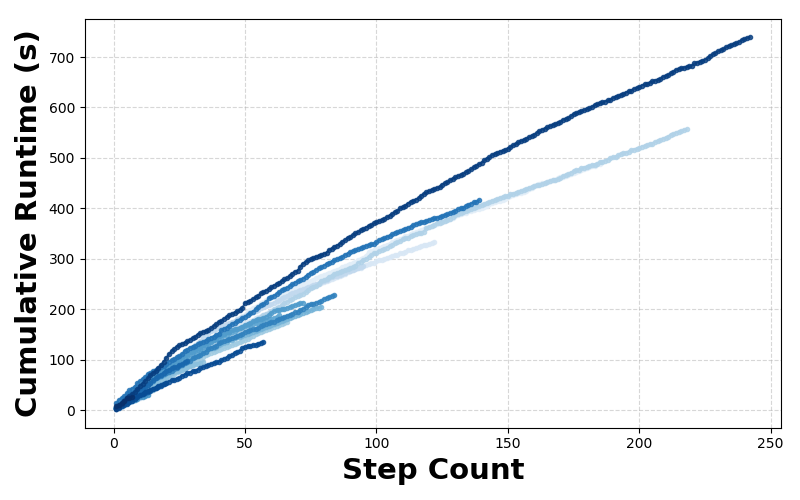}
		\caption{Cumulative runtime across steps of the 20 sampled traces.}
		\label{fig:runtimesteps}
	\end{minipage}%
	\hspace{0.02\textwidth}
	\begin{minipage}[b]{0.48\textwidth}
		\centering
		\includegraphics[height=4cm]{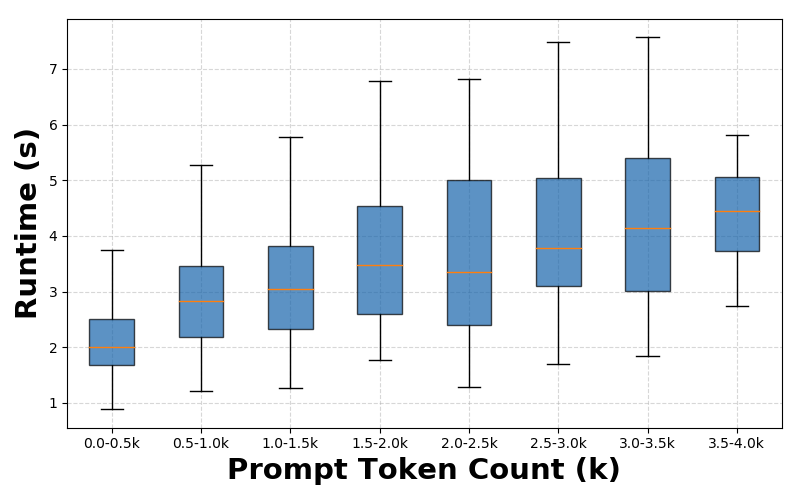}
		\caption{Time across prompt token counts of the 20 sampled traces.  (x-axis truncated at 4k)}
		\label{fig:runtimeprompt}
	\end{minipage}
\end{figure}

In this section, we evaluate the runtime overhead of {\toolname}.
We profile {\toolname}'s runtime over 20 trajectories drawn from agent frameworks in \datasetname{} as well as the Who\&When dataset.
All measurements use GPT-4o as the base model.
Figure~\ref{fig:runtimesteps} shows cumulative runtime per step, with each line representing one of the 20 sampled trajectories. 
The per-trace runtime overhead ranges from $\sim 25\,\mathrm{s}$ (a 4-step trace with 7 calls) to $\sim 750\,\mathrm{s}$ (a 120-step trace with 242 calls).
Figure~\ref{fig:runtimeprompt} shows a box plot of runtime across all trajectories, grouped by the number of prompt tokens. 
The plot is truncated at 4k tokens to focus on the majority of cases; a small number of outliers with higher token counts and runtimes are omitted for clarity.
These results indicate that {\toolname} introduces a manageable runtime overhead, with latency spikes occurring only in rare cases.

\begin{figure}[H]
	\centering
	\includegraphics[width=\linewidth]{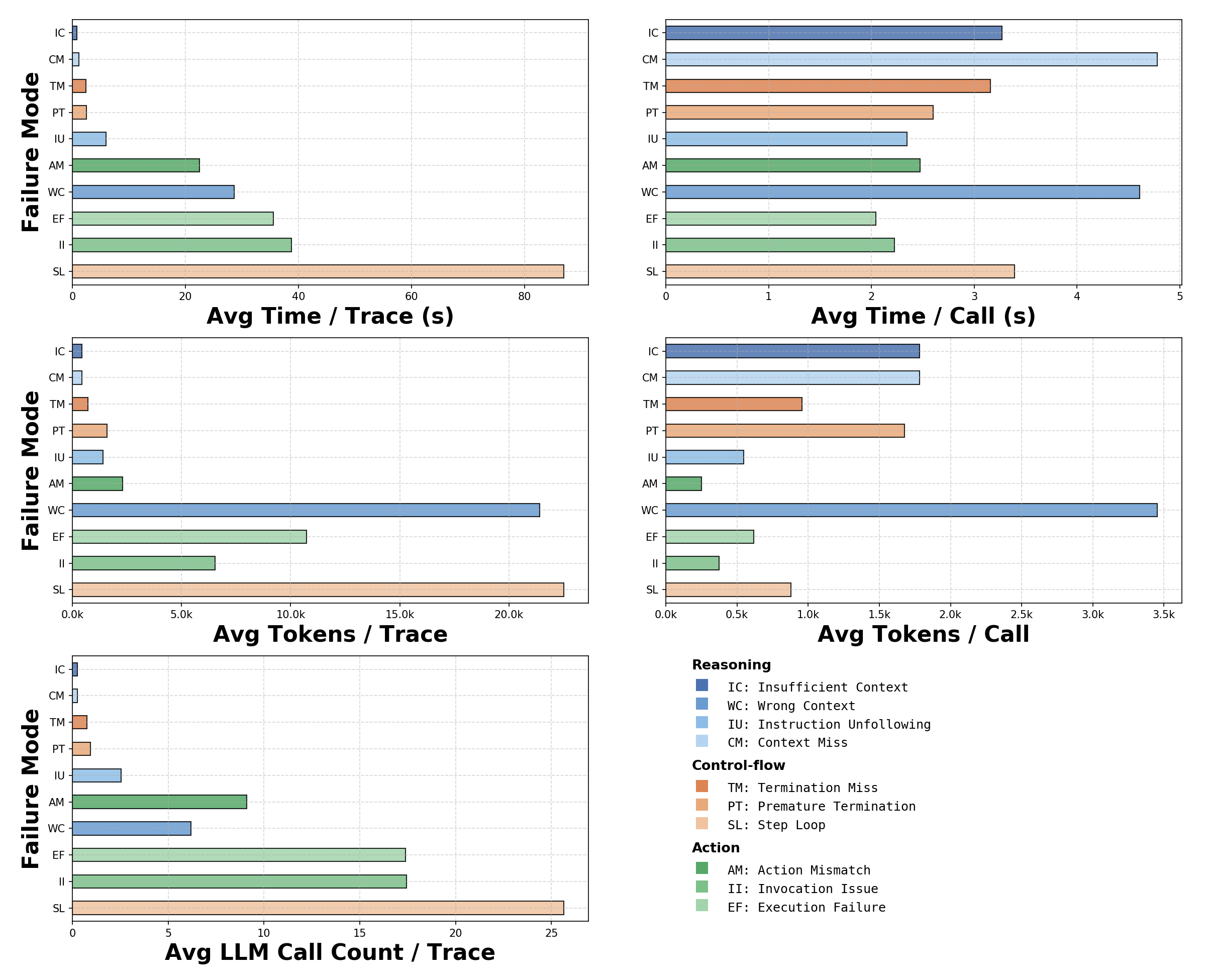}
	\caption{Time and LLM calls per invariant check across 20 sampled traces.}
	\label{fig:runtimertotal}
\end{figure}

Figure~\ref{fig:runtimertotal} decomposes the runtime cost by {\failname} taxonomy.
We observe that the \textit{Step Loop} accounts for the largest runtime ($\sim 85\,\mathrm{s}$). 
This is expected, as detecting step loops {\failname} requires pairwise segment comparisons, causing the number of LLM calls to scale quadratically with the number of loop-candidate windows. 
The next most time-consuming {\failname} are \textit{Invocation Issue} and \textit{Execution Failure}. 
Each invokes $\sim 17$ calls per trace, but incurs only $\sim 2\,\mathrm{s}$ per call due to their relatively compact prompts ($\sim 375$ and $\sim 600$ tokens on average, respectively). 
\textit{Wrong Context} has the highest per-call cost because its judge function processes full step context and neighboring annotations. 
However, it is invoked less frequently ($\sim 6$ calls per trace), keeping its total runtime moderate ($\sim 30\,\mathrm{s}$ per trace).

In general, we can conclude that the average LLM call count has a larger impact on runtime than the average prompt token count, as the time per call remains relatively stable across different {\failname} checks.
This implies that optimizing the number of calls---e.g., by merging multiple calls into a single call---may yield more significant runtime reductions than simply optimizing prompt length. 
Moreover, since the failure checks are independent of each other, they can be executed in parallel. 
Additionally, calls with different prompt lengths can be packed by type, further improving throughput and reducing total runtime. If cacheable prefix information is available, calls with the same prefix can be prioritized for packing, allowing subsequent requests to reuse cached key-value states and further accelerate execution.
Overall, these observations suggest several potential runtime optimizations that could be applied to further improve efficiency.


\section{Related Work}\label{sec:related}

\textbf{Failures of Agents.} As LLM-based agents tackle increasingly complex tasks such as planning, web navigation, and tool use, understanding their failure modes becomes critical. Prior work highlights issues like tool misuse, hallucinated goals, and premature task termination. Architectural improvements such as memory modules or hierarchical task decomposition (e.g., Voyager~\citep{wang2023voyager}, AssistGPT~\citep{gao2023assistgpt}) primarily focus on improving task success, offering limited introspection into why failures occur. 

Existing agent failure studies~\citep{cemri2025multi} classify overall agent mistakes and construct failure taxonomies. In particular, \citet{zhang2025agent, zhu2025raffles} identify failure steps using LLMs as judges or via LLM reasoning, whereas \citet{zhang2025agentracer} fine-tunes models to automatically detect agent failures.
Differently, {{\toolname}} introduces a structured, interpretable, and reliable neuro-symbolic framework that encodes agent execution trajectories into \emph{behavioral abstractions}. By representing raw trajectories in a structured form, {\toolname} enables neural-invariant-guided reasoning to systematically analyze agent behavior against formally defined invariants, pinpoint the exact step at which a failure occurs, and classify its type.
Since {\toolname} is built on LLMs as the base models, its work is orthogonal to that of fine-tuned detection models. On the other hand, the structured analysis and design of invariants here are orthogonal to directly using LLMs as judges, and can thus benefit existing approaches.

\textbf{LLM as a Judge.} Recent works~\citep{gu2024survey, li2024generation} explore using LLMs as evaluators to assess output quality, correctness, or alignment. Benchmarks like Arena~\citep{chiang2024chatbot} rank responses across tasks including QA, dialogue, and reasoning. While LLMs demonstrate meta-evaluative abilities, critiques~\citep{chen2024humans, szymanski2025limitations} show they can be unreliable or inconsistent, especially for nuanced reasoning or domain-specific instructions. Compared to LLM-as-judge approaches, {\toolname} leverages structured trajectory representations to guide step-wise reasoning diagnosis, offering interpretable and precise failure identification rather than holistic or subjective black-box assessments.

\section{Conclusion}\label{sec:conclusion}

We presented \toolname{}, a neuro-symbolic framework for diagnosing failures in LLM agents that abstracts long and heterogeneous agent trajectories into {\graphabbr}s, applies neural invariant checking to surface candidate {\failname}s, and selects the decisive error that best explains the final degraded outcome, enabling more localized, auditable, and interpretable diagnoses than monolithic LLM-as-a-judge approaches. Across both the public Who\&When benchmark and our new benchmark \datasetname{}, \toolname{} achieves substantially stronger failure localization and attribution performance than prior baselines, while also supporting fine-grained classification of {\failname}s and suggesting that structured behavioral abstraction and invariant-guided reasoning are promising foundations for debugging increasingly complex agent systems.

\clearpage

\bibliographystyle{iclr-bib}
\bibliography{main}

\clearpage

\appendix
\appendix

\section{Neural Invariant Violations}
\label{app:invar}

In this section, we present the definition of neural invariants and describe how violations of these invariants manifest in agent trajectories.

\subsection{Wrong Context} 

Wrong context occurs when an agent uses irrelevant, fabricated, or improperly
assumed information in its reasoning—either failing to clarify genuinely
ambiguous input or hallucinating prior context that does not exist.

Let $\mathbf{N}_{output} \subseteq \mathbf{N}$ be the set of output nodes and
let $\mathbf{N}_{user} \subseteq \mathbf{N}$ be the set of user-input nodes.
For the candidate node $n_w$ and its backward context windows, define two
judging functions.
$\texttt{failedToClarify}(user_t, output_w)$ evaluates whether the user input
is ambiguous but the assistant proceeded without asking for clarification.
$\texttt{hallucinated}(output_w, W_1, \dots, W_m)$ evaluates whether the
assistant asserts prior context or tool results that do not appear in the
checked context windows $W_1, \dots, W_m$.

The invariant violation is expressed as:
\begin{multline*}
    iv_{wrong} ::= \exists~n_w \in \mathbf{N}_{output} : \\
    \texttt{failedToClarify}(user_t,\; output_w) ~\vee~
    \texttt{hallucinated}(output_w,\; W_1, \dots, W_m).
\end{multline*}

Violation indicates that the agent is reasoning from an unsupported or
improperly clarified contextual state, potentially leading to downstream errors.
If an LLM is used as the judge, the threshold is replaced by a binary classification label.

\subsection{Instruction Unfollowing}

Instruction unfollowing occurs when the agent's output fails to
satisfy the given instructions, including role requirements, formatting
constraints, and requested behavior.

Let $\mathbf{N}_{output} \subseteq \mathbf{N}$ be the set of nodes representing
final outputs, and let $\mathbf{I}$ be the set of extracted instructions
from system or instruction-bearing steps. For an output node $n_f \in \mathbf{N}_{output}$
and an instruction $i \in \mathbf{I}$, define a judging function
$\texttt{followed}(instruction_i, output_f)$ that evaluates whether the output
satisfies the instruction's requirements.

The invariant violation is expressed as:
\[
    iv_{unfollowed} ::= \exists~i \in \mathbf{I},~n_f \in \mathbf{N}_{output} :
    \neg\, \texttt{followed}(instruction_i,\; output_f).
\]

Violation indicates that the agent's final output does not satisfy the
instructions governing that output--for example through role violation,
formatting violation, missing sections, ignored instructions, or refusal
without valid reason.
If an LLM is used as the judge of $\texttt{followed}$, the threshold is replaced by a binary classification label.

\subsection{Insufficient Context}

Insufficient context occurs when an agent explicitly claims that necessary
information is missing, and the claimed information is indeed absent from the
available context.

Let $\mathbf{N}_{claim} \subseteq \mathbf{N}$ be the set of nodes in the \graphabbr{} where the
assistant claims missing or insufficient context (detected via textual hooks
such as ``cannot be determined'', ``insufficient context'', or explicit
\texttt{insufficient\_context\_flag} annotations). For a candidate node
$n_c \in \mathbf{N}_{claim}$ and its backward context windows
$W_1, \dots, W_m$, define a judging function
$\texttt{infoAbsent}(claim_c, W_1, \dots, W_m)$ that evaluates whether the
claimed missing information is genuinely absent from all checked context windows.

The invariant violation is expressed as:
\[
iv_{insuff} ::= \exists~n_c \in \mathbf{N}_{claim} :
\texttt{infoAbsent}(claim_c,\; W_1, \dots, W_m).
\]

Violation indicates that the agent's lack-of-context claim is grounded in
genuinely missing evidence—the required information truly does not exist in
the provided context.
If an LLM is used as the judge of $\texttt{infoAbsent}$, the threshold is replaced by a binary classification label.

\subsection{Context Miss}

Context miss occurs when the agent claims information is missing even
though it is available in the prior context, or when the agent failed to utilize available context when it was needed for reasoning or action execution.

Using the same candidate node set $\mathbf{N}_{claim}$ as the invariant of Insufficient Context, for a candidate node
$n_c \in \mathbf{N}_{claim}$ and its backward context windows
$W_1, \dots, W_m$, define a judging function
$\texttt{infoPresent}(claim_c, W_1, \dots, W_m)$ that evaluates whether the
claimed missing information actually exists in at least one context window, and
a function $\texttt{confidentUngrounded}(output_c, W_1, \dots, W_m)$ that
evaluates whether the assistant reached a definitive conclusion without citing
supporting evidence from the context.

The invariant violation is expressed as:
\begin{multline*}
	iv_{ctxmiss} ::= \exists~n_c \in \mathbf{N}_{claim} :\\
	\texttt{infoPresent}(claim_c,\; W_1, \dots, W_m) ~\vee~
	\texttt{confidentUngrounded}(output_c,\; W_1, \dots, W_m).
\end{multline*}

Violation indicates that the agent mishandled available or required context
rather than merely lacking it -- either by ignoring present information or by
asserting conclusions without grounding.
If an LLM is used as the judge, the threshold is replaced by a binary classification label.

\subsection{Termination Miss}

Termination miss occurs when the assistant already produced the final answer
but continued to act (reasoning or tool calls) instead of stopping.

Let $\mathbf{N}_{output} \subseteq \mathbf{N}$ be the set of nodes representing
final outputs, and let $\mathbf{N}_{post} \subseteq \mathbf{N}$ be the set of
assistant nodes occurring after the final output step.
Define a judging function $\texttt{continued}(n_f, \mathbf{N}_{post})$ that
evaluates whether there is any post-final-output assistant activity
(tool calls, reasoning, or mixed actions).

The invariant violation is expressed as:
\[
    iv_{termmiss} ::= \exists~n_f \in \mathbf{N}_{output} :
    |\mathbf{N}_{post}| > 0 ~\wedge~
    \texttt{continued}(n_f,\; \mathbf{N}_{post}).
\]

Violation indicates that the agent failed to recognize task completion after
producing its final answer, leading to redundant or infinite reasoning steps.
If an LLM is used as the judge of $\texttt{continued}$, the threshold is replaced by a binary classification label.

\subsection{Premature Termination}

Premature termination occurs when the agent stops reasoning or task execution
before satisfying the required end conditions.

Let $n_{last}$ denote the last executed node in the \graphabbr{}, and $H$ denote the entire conversation history. Define $\texttt{endReq}(H)$ as the end-condition requirements extracted from the first
step carrying a non-empty \texttt{end\_condition\_instruction} annotation, together with
required format labels that the final output may need to satisfy(e.g.\ \texttt{Explanation}, \texttt{Exact Answer}, \texttt{Confidence}). Define a judging function
$\texttt{goalMet}(output_{last}, endReq)$ that evaluates whether the final output
satisfies the required end-condition content.

The invariant violation is expressed as:
\[
    iv_{premature} ::= \neg\, \texttt{goalMet}(output_{last},\; \texttt{endReq}(H)).
\]

Violation indicates that the agent terminated before producing the required
output content, resulting in incomplete solutions or missed requirements.
If an LLM is used as the judge of $\texttt{goalMet}$, the threshold is replaced by a binary classification label.

\subsection{Step Loop}

Step loop occurs when a consecutive group of steps repeats the same
action or intent without advancing the task.

Let $\mathbf{N}$ be the set of nodes in the \graphabbr{}. For a loop window
of size $k$ and a start index $t$, define two consecutive segments
$A_t = (n_{t-k}, \dots, n_{t-1})$ and $B_t = (n_t, \dots, n_{t+k-1})$.
Define a judging function $\texttt{loopRepeat}(A_t, B_t)$ that evaluates
whether segment $B_t$ semantically repeats segment $A_t$ with no observable
progress—by comparing roles, step purposes, actions, reasoning summaries, and
tool outcomes across corresponding positions.

The invariant violation is expressed as:
\[
    iv_{loop} ::= \exists~t :
    \texttt{loopRepeat}(A_t,\; B_t).
\]

Violation indicates that the agent is trapped in repeated local behavior
without moving the task forward.
If an LLM is used as the judge of $\texttt{loopRepeat}$, the threshold of semantic similarity is replaced by a binary classification label.

\subsection{Action Mismatch}

Action mismatch occurs when the assistant's executed action contradicts or
does not align with the intent expressed in the immediately preceding reasoning step.

Let $\mathbf{N}_{action} \subseteq \mathbf{N}$ be the set of nodes representing
action steps (i.e.\ tool calls or output steps), and let $\mathbf{N}_{reason} \subseteq \mathbf{N}$
be the set of reasoning nodes that establish step purposes.
For an action node $n_t$ and its preceding reasoning node $n_{t-1}$, define a
judging function
$\texttt{aligned}(purpose_{t-1}, action_t, tool\_name_t, tool\_args_t)$
that evaluates whether the action type matches the step intent, the tool choice
aligns with task semantics, and the action advances the task.

The invariant violation is expressed as:
\begin{multline*}
    iv_{mismatch} ::= \exists~n_t \in \mathbf{N}_{action},~n_{t-1} \in \mathbf{N}_{reason} : \\
    \text{\(\neg\, \texttt{aligned}(purpose_{t-1},\; action_t,\; tool\_name_t,\; tool\_args_t)\)}\\
\end{multline*}

Violation indicates that the agent took an action that is not well aligned with
its stated local purpose -- for example through tool misuse, a missing action,
action substitution, or hallucinated progress.
If an LLM is used as the judge of $\texttt{aligned}$, the threshold is replaced by a binary classification label.

\subsection{Invocation Issue}

Invocation issue occurs when the assistant's tool call does not comply with the
tool's specification—missing required parameters, invalid types, unknown actions,
or calling a tool not found in the provided tool list.

Let $\mathbf{N}_{call} \subseteq \mathbf{N}$ be the set of nodes representing
tool invocations. For a node $n_t \in \mathbf{N}_{call}$, let $spec_t$ denote
the matching tool schema from the tool list. Define a judging function
$\texttt{schemaOK}(call_t, spec_t, response_t)$ that evaluates whether the
invocation complies with the tool's parameter specification and dependencies.

The invariant violation is expressed as:
\[
    iv_{invocation} ::= \exists~n_t \in \mathbf{N}_{call} :
    \neg\, \texttt{schemaOK}(call_t,\; spec_t,\; response_t).
\]

Violation indicates that the tool invocation is not valid under the interface
expected by the environment -- for example through a missing required parameter,
an invalid parameter type, an unknown action, incompatible parameter combinations,
a tool not found in the tool list, or a tool-reported invocation error.
If an LLM is used as the judge of $\texttt{schemaOK}$, the threshold is replaced by a binary classification label.

\subsection{Execution Failure}

Execution failure occurs when the tool's response indicates that the requested
operation did not complete successfully (e.g.\ exception, HTTP error, empty
result, timeout, or permission denial).

Let $\mathbf{N}_{tool} \subseteq \mathbf{N}$ be the set of nodes associated
with tool responses. For a node $n_t \in \mathbf{N}_{tool}$, define a judging
function $\texttt{execFailed}(response_t)$ that evaluates whether the tool
response indicates an error or non-success.

The invariant violation is expressed as:
\[
    iv_{exec} ::= \exists~n_t \in \mathbf{N}_{tool} :
    \texttt{execFailed}(response_t).
\]

Violation indicates that the requested operation did not execute as intended,
potentially due to tool errors, empty results, timeouts, permission denial,
rate limiting, invalid input, or network issues.
If an LLM is used as the judge of $\texttt{execFailed}$, the threshold is replaced by a binary classification label.

\label{app:annotation}

\section{{\failnameup} Distribution of {\datasetname}}\label{app:datadis}

\definecolor{reaColor}{RGB}{210,236,165}
\definecolor{ctrlColor}{RGB}{62,167,196}
\definecolor{actColor}{RGB}{44,56,145}
\definecolor{icColor}{RGB}{222,242,188}
\definecolor{wcColor}{RGB}{194,228,149}
\definecolor{iuColor}{RGB}{155,207,111}
\definecolor{cmColor}{RGB}{109,171,75}
\definecolor{tmColor}{RGB}{138,219,235}
\definecolor{ptColor}{RGB}{74,187,213}
\definecolor{slColor}{RGB}{28,130,176}
\definecolor{amColor}{RGB}{106,133,213}
\definecolor{iiColor}{RGB}{63,88,179}
\definecolor{efColor}{RGB}{34,45,122}

\newcommand{\piesector}[6]{%
    \path[draw=white, line width=0.9pt, fill=#4]
        (0,0) -- (#1:#3) arc[start angle=#1,end angle=#2,radius=#3] -- cycle;
    \pgfmathsetmacro{\midangle}{(#1 + #2) / 2}
    \node[align=center,font=#6] at (\midangle:{0.63*#3}) {#5};
}

\begin{figure}[H]
    \centering
    \begin{minipage}[c]{0.62\linewidth}
        \centering
        \begin{tikzpicture}[scale=0.98]
            \def\r{2.9}
            \piesector{140.0}{272.6}{\r}{reaColor}{Rea\\36.0\%}{\small\bfseries}
            \piesector{272.6}{372.3}{\r}{ctrlColor}{Ctrl\\31.0\%}{\small\bfseries}
            \piesector{372.3}{500.0}{\r}{actColor}{Act\\36.0\%}{\small\bfseries}
        \end{tikzpicture}
    \end{minipage}%
    \begin{minipage}[c]{0.28\linewidth}
        \raggedright
        \small
        \begin{tabular}{l l}
            \textcolor{reaColor}{\rule{10pt}{10pt}} & Rea (Reasoning) \\
            \textcolor{ctrlColor}{\rule{10pt}{10pt}} & Ctrl (Control-flow) \\
            \textcolor{actColor}{\rule{10pt}{10pt}} & Act (Action) \\
        \end{tabular}
    \end{minipage}
    \caption{Failure Category Distribution of \datasetname}
    \label{fig:gtcatpie}
\end{figure}
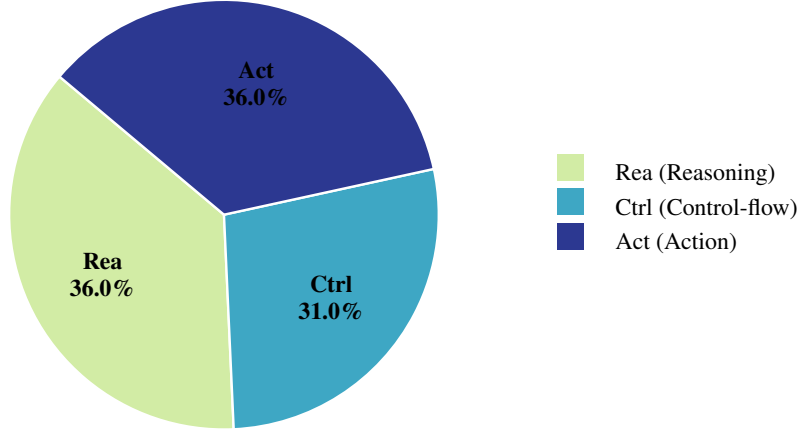
\begin{figure}[H]
    \centering
    \begin{minipage}[c]{0.60\linewidth}
        \centering
        \begin{tikzpicture}[scale=0.95]
            \def\r{2.8}
            \piesector{90.0}{123.9}{\r}{icColor}{IC\\9.6\%}{\scriptsize\bfseries}
            \piesector{123.9}{153.2}{\r}{wcColor}{WC\\8.3\%}{\scriptsize\bfseries}
            \piesector{153.2}{193.1}{\r}{iuColor}{IU\\9.9\%}{\scriptsize\bfseries}
            \piesector{193.1}{222.4}{\r}{cmColor}{CM\\5.3\%}{\scriptsize\bfseries}
            \piesector{222.4}{256.3}{\r}{tmColor}{TM\\9.6\%}{\scriptsize\bfseries}
            \piesector{256.3}{290.3}{\r}{ptColor}{PT\\12.5\%}{\scriptsize\bfseries}
            \piesector{290.3}{322.0}{\r}{slColor}{SL\\8.9\%}{\scriptsize\bfseries}
            \piesector{322.0}{368.9}{\r}{amColor}{AM\\13.2\%}{\scriptsize\bfseries}
            \piesector{368.9}{397.0}{\r}{iiColor}{II\\7.9\%}{\scriptsize\bfseries}
            \piesector{397.0}{450.0}{\r}{efColor}{EF\\14.9\%}{\scriptsize\bfseries}
        \end{tikzpicture}
    \end{minipage}%
    \begin{minipage}[c]{0.28\linewidth}
        \raggedright
        \small
        \begin{tabular}{l l l@{\hspace{1.2em}}l l}
            \textcolor{icColor}{\rule{9pt}{9pt}} & IC &
            \textcolor{ptColor}{\rule{9pt}{9pt}} & PT \\
            \textcolor{wcColor}{\rule{9pt}{9pt}} & WC &
            \textcolor{slColor}{\rule{9pt}{9pt}} & SL \\
            \textcolor{iuColor}{\rule{9pt}{9pt}} & IU &
            \textcolor{amColor}{\rule{9pt}{9pt}} & AM \\
            \textcolor{cmColor}{\rule{9pt}{9pt}} & CM &
            \textcolor{iiColor}{\rule{9pt}{9pt}} & II \\
            \textcolor{tmColor}{\rule{9pt}{9pt}} & TM &
            \textcolor{efColor}{\rule{9pt}{9pt}} & EF
        \end{tabular}
    \end{minipage}
    \caption{Failure Mode Distribution of \datasetname}
    \label{fig:gtpie}
\end{figure}
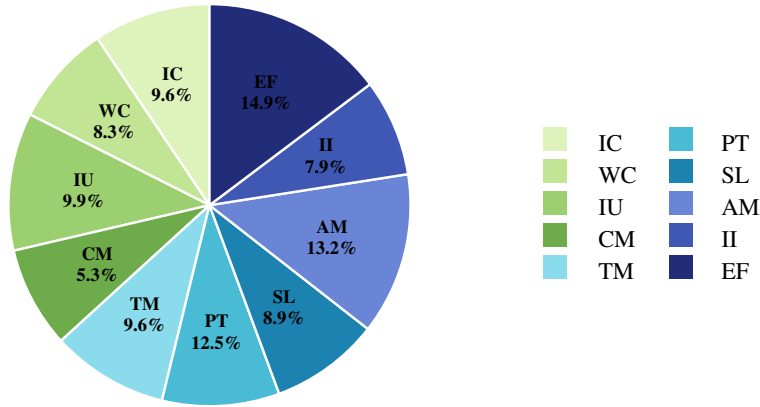

This section provides a detailed breakdown of the failure composition in \datasetname{}, aiming to characterize the structural diversity of diagnostic targets. Understanding the distribution of failure types is crucial, as it directly reflects the heterogeneity and complexity of trajectory-level errors, which in turn determines the difficulty of localization and attribution.

As shown in Figure~\ref{fig:gtcatpie}, failures are broadly categorized into three high-level groups: reasoning-related (Rea), control-flow-related (Ctrl), and action-related (Act) failures. The distribution is relatively balanced across these categories, indicating that no single failure type dominates the dataset. This balance ensures that evaluation is not biased toward a specific reasoning pattern, but instead requires robust handling of diverse failure mechanisms spanning cognitive reasoning, procedural execution, and decision-level actions.

Beyond coarse-grained categorization, Figure~\ref{fig:gtpie} further decomposes failures into fine-grained modes, including IC, WC, IU, CM, TM, PT, SL, AM, II, and EF, with their full names listed in Table~\ref{tab:error-main}.
It is worth noting that the slight distributional variations observed in the dataset stem from the construction process, in which we inject failures into successfully completed trajectories. During this fault injection process, the LLM is required to judge whether the injected {\failname} is consistent with the context of the original trajectory, which introduces subtle biases into the resulting failure distribution.
Such a distributional property is important for evaluating diagnostic systems. In particular, it prevents models from overfitting to a narrow subset of failure patterns and instead encourages more generalizable failure localization and attribution capabilities.
Overall, the failure distributions in \datasetname{} demonstrate both semantic diversity and structural balance, making it a suitable benchmark for evaluating robust trajectory diagnosis methods under realistic and varied failure conditions.

\section{Case Study of Who\&When Dataset}\label{whonwhen}

Traces in the Who\&When dataset often contain multiple errors of varying types and severities, rather than a single decisive mistake. The annotated “mistake step” typically marks the earliest detectable issue in a trajectory, but it does not necessarily correspond to the most causally critical failure.

Consider the following example~\cite{whonwhen2}. The ground-truth mistake is annotated at Step 4, where the WebSurfer retrieves only a search results page without extracting structured information (e.g., series list, number of seasons, or ratings). While this step reflects information insufficiency, it still represents a plausible intermediate exploration step and does not inherently prevent successful task completion. This interpretation is further supported by the system's internal state, which indicates continued progress at this stage.

In contrast, later steps (e.g., Steps 11 and 15) exhibit a more critical failure mode. The WebSurfer repeatedly performs the same action---clicking the same page and observing an identical viewport---without any meaningful state update or information gain. This behavior forms a stagnation loop, which is explicitly detected by the system (is\_in\_loop: true) and directly blocks further progress toward task completion. Notably, even after the Orchestrator issues refined instructions, the WebSurfer continues to repeat the same ineffective actions, indicating a failure to recover from the erroneous state.

Finally, the trajectory does not terminate due to successful completion or reasoning convergence. Instead, it is interrupted by a system-level failure during the orchestration phase. Specifically, a content filtering violation triggers a ResponsibleAI policy error, which results in a BadRequestError during the orchestrator’s model client call for ledger updates. Consequently, execution halts prematurely.

This example illustrates that multiple failure modes can coexist within a single trajectory, spanning early-stage information insufficiency and later-stage behavioral stagnation. Crucially, the annotated mistake corresponds to the former, rather than the latter, which is more decisive for the final outcome.
This limitation may reduce its effectiveness in evaluating a model’s ability to localize the most impactful failure points. More broadly, effective failure localization requires not only identifying the first erroneous step, but also reasoning about its downstream impact on task completion.

\end{document}